# Formal Language Knowledge Corpus for Retrieval Augmented Generation

Majd Zayyad, Yossi Adi

The Hebrew University of Jerusalem

*Abstract-* The integration of retrieval-augmented techniques with LLMs has shown promise in improving performance across various domains. However, their utility in tasks requiring advanced reasoning, such as generating and evaluating mathematical statements and proofs, remains underexplored. This study explores the use of Lean, a programming language for writing mathematical proofs, to populate the knowledge corpus used by RAG systems. We hope for this to lay the foundation to exploring different methods of using RAGs to improve the performance of LLMs in advanced logical reasoning tasks.

## I. INTRODUCTION

Large Language Models (LLMs) have demonstrated remarkable capabilities across a wide range of tasks, but they still face significant challenges, particularly in generating accurate and reliable information. One of the key issues is their tendency to produce hallucinated or incorrect responses. This has led to the use of Retrieval-Augmented Generation (RAGs) [Gao et al., 2023, Mialon et al., 2023] in an effort to overcome such challenges, since RAGs allow the models to rely on verified external sources of information, which can offer increased accuracy to the generated data and combat hallucinations. However, RAGs still fail to ground LLMs when generating solutions to logical questions, and LLMs still fall short when attempting tasks that require the use of general reasoning skills. This manifests especially when it comes to mathematical reasoning.

One key issue is the difficulty LLMs face in achieving semantic understanding and contextual reasoning in mathematical language, often leading to incorrect or incomplete formalization of mathematical concepts [Ying et al., 2024]. This gap arises because mathematical language requires a depth of concept comprehension that is challenging to encode in token-based LLM frameworks [Gao et al., 2024]. LLMs also struggle with long-term dependencies inherent in mathematical reasoning, as solutions to mathematical problems often rely on concepts and steps from earlier sections of a text [Lin et al., 2024]. Given the models' limited memory, they have difficulty establishing continuity across extended logical arguments. Precision is another area of difficulty, as the inherent ambiguity in natural language (NL) can lead LLMs to make unintended assumptions, which do not align with the rigor required by mathematical languages [Gao et al., 2024].

Maintaining logical coherence across multiple logical steps is a further challenge; models often create disjointed or incoherent solutions, reflecting an inability to understand the sequence and relationships required in proofs [Ying et al., 2024]. Handling mathematical symbols and complex notations also poses issues since these elements are often misinterpreted by LLMs trained primarily on text rather than on specialized mathematical symbols [Agrawal et al., 2022]. Recursive processes present additional complexity, as they require models to handle multi-layered logical states and iterative reasoning, which LLM architectures are not well-suited to [Xin et al., 2024].

The study will investigate the use of formalized mathematical statements in Lean, to build a knowledge corpus for the use by RAGs. This approach involves translating NL queries into formal language (FL), in order to query and represent data, potentially improving the performance of LLMs in a math-focused question-and-answer (QnA) application. Additionally, we will evaluate this method against established benchmarks, such as the Mathematics Dataset developed by Google [Saxton et al., 2019], in order to gauge its effectiveness compared to traditional RAG configurations that represent and retrieve information in NL. This research aims to assess whether integrating FL in the RAG process can yield advantages over conventional NL-based approaches.

## II. BACKGROUND

*A. Retrieval-Augmented Language Models*

RAG is a sophisticated framework designed to enhance language models by coupling them with external retrieval systems, addressing limitations inherent in static, solely parameter-based language models. RAG integrates a dual-component architecture where a retriever dynamically searches a structured external corpus for relevant information based on the input query, and a generator LLM uses the retrieved content as context to generate accurate and contextually enriched responses [Gao et al., 2023, Mialon et al., 2023]. This setup mitigates common issues such as hallucinations and factual inaccuracies in language models by grounding generated text in real-world, verified information. In practice, RAG systems employ dense vector embeddings to ensure retrieval relevance, capturing semantic relationships within documents beyond mere keyword matching. The retrieved information is subsequently fed into the generator, allowing it to synthesize data with pre-existing knowledge for enhanced coherence and contextual accuracy.

RAGs can employ two main types of retrieval mechanisms: dense and sparse [Mialon et al., 2023]. Sparse retrievers rely on bag-of-words representations, excelling at finding documents with high term overlap to the query, while dense retrievers utilize neural network embeddings to capture semantic similarities, enhancing the model's comprehension of related concepts. By appending retrieved documents directly to the model's context, these retrievers allow the language model to ground its responses in a broader context, thereby increasing accuracy and factual consistency across complex tasks.

The success of retrieval-augmented models in various domains has catalyzed interest in their application to more demanding reasoning tasks. A great representative of such tasks is the construction and verification of mathematical proofs, which requires solving problems step-by-step, and generating precise mathematical statements. Recent approaches, such as chain-of-thought (CoT) prompting [Lewkowycz et al., 2022] combined with retrieval, highlight the potential for retrieval-augmented models to provide sequential reasoning support. These models can generate reasoning paths interspersed with retrieval steps to guide complex problem-solving processes, such as multi-step question answering, enabling models to leverage external information dynamically at each reasoning stage.

*B. Autoformalization*

Autoformalization is the process of translating informal mathematical expressions, typically written in NL or standard mathematical notation, into formalized, machine-readable language that theorem provers and proof assistants (such as Lean) can interpret and verify [Wu et al. (2022)]. This transformation process is complex, requiring the formal system to not only translate symbols accurately but also to grasp the semantic and logical nuances of mathematical language. The goal of autoformalization is to enable computers to autonomously produce valid formal statements from human-readable text, thereby reducing the time and expertise needed to encode informal statements manually into formal systems like Lean.

LLMs can perform autoformalization by using Few-shot prompting. Few-shot prompting is a strategy that provides LLMs with small sets of example pairs, illustrating how informal language maps to formalized statements. These examples help guide the model in recognizing the patterns and syntactic structures unique to formal mathematical language. The LLM can then apply this learned structure to new informal inputs, generating outputs that are syntactically and semantically aligned with formal systems' requirements. This approach leverages the model's capacity to generalize from limited examples, enabling it to interpret complex mathematical statements and produce formal representations accurately.

The formalization process requires the LLM to maintain logical coherence across multi-step arguments and correctly interpret mathematical abstractions—tasks that are beyond the capabilities of many general-purpose language models. However, LLMs that are fine-tuned with carefully curated formalization examples show promising results, achieving increasingly accurate interpretations of informal mathematical language. This advancement has significant implications for fields such as formal verification and automated reasoning [Wu et al., 2022].

The recent advancements in autoformalization and automated theorem proving build upon foundational datasets and innovative training techniques to bridge the gap between NL and formal mathematical systems. In projects such as the Lean Workbook [Ying et al., 2024], large-scale datasets have been compiled to convert informal mathematical language, including word problems and theorems expressed in standard notation, into structured, formal syntax for theorem provers. These datasets encompass a wide range of mathematical problems, creating resources that train models to accurately interpret, translate, and formalize complex mathematical statements. The availability of such structured data enables machine learning models to address the complexities inherent in formalizing mathematical expressions.

Following this groundwork, synthetic datasets specifically designed for LLM training in theorem proving, such as those used in DeepSeek-Prover [Xin et al., 2024], have been developed to refine models' capabilities in searching and verifying proofs. These datasets employ reinforcement learning and Monte Carlo Tree Search to optimize proof searches, training models to interact with theorem provers and receive real-time feedback. This iterative training enables models to handle sophisticated mathematical logic, improving their performance in generating and verifying proofs autonomously.

## III. Implementation

In this section, we detail the architecture and implementation of the two RAGs developed in this study: the NL RAG and the FL RAG. Both systems are designed to enhance the mathematical reasoning capabilities of LLMs by integrating external knowledge bases. The key difference lies in how the query is processed and how relevant documents are retrieved from the knowledge corpus.

### A. Natural Language RAG

The NL RAG system leverages a corpus of mathematical problems and solutions expressed in NL. The architecture consists of several sequential components that work together to process the user's query and generate an accurate response. Initially, the user inputs a mathematical question in NL. This query is then converted into a high-dimensional vector representation using a pre-trained embedding model, specifically OpenAI's text-embeddings-ada-002. This embedding captures the semantic meaning of the query and facilitates similarity comparisons with other text embeddings.

Subsequently, the query embedding is used to perform a similarity search within a vector database containing embeddings of NL mathematical statements and solutions. The vector database was constructed from datasets such as the MATH dataset [Hendrycks et al., 2021], which includes high school-level math problems. The system retrieves the top $k$ most relevant documents based on cosine similarity between the query embedding and the document embeddings. These retrieved documents are then concatenated to form a context, which is combined with the original query to provide additional information for the language model. Finally, the combined context and query are input into a GPT-4o model, the system prompt for the model is *"Solve the following math problem. Give the final answer within {}. Like so: the final answer is {answer}"*. The model generates an answer by leveraging both the retrieved information and its pre-trained knowledge. The generated response is then presented to the user.

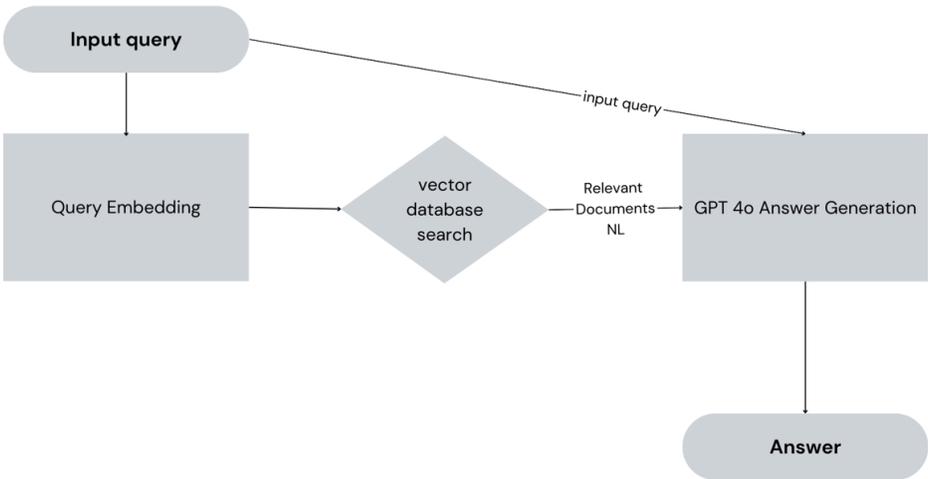

Natural Language RAG Flowchart

*B. Formal Language RAG*

The FL RAG system incorporates formal mathematical language, specifically Lean code, into the retrieval process. This approach involves an additional step of translating the NL query into a formal representation before performing retrieval. When the user inputs a mathematical question in NL, the system first translates this query into a Lean formal statement. This translation is accomplished using a fine-tuned version of GPT-4o-mini, which was trained on a dataset [Ying et al., 2024] consisting of pairs of NL statements and their corresponding Lean formalizations.

The Lean-formatted query is then embedded using text-embeddings-ada-002 (we found text-embeddings-ada-002 to be capable of handling FL as well as text). Next, the query embedding is used to search within a vector database constructed from formal mathematical statements in Lean. This database was created by translating the same MATH dataset [Hendrycks et al., 2021], in order to ensure results aren't impacted by the knowledge corpus used.

The system retrieves the top *k* Lean documents that are semantically similar to the query. These retrieved documents are then combined with the original NL query to form the context for the language model. The combined context and query are input into the GPT-4o model, which generates an answer that incorporates formal reasoning, the system prompt for the model was the same as the one used in the NL RAG, which is *"Solve the following math problem. Give the final answer within {}. Like so: the final answer is {answer}"*. The GPT-4o model is prompted with FL context. The generated answer is then presented to the user.

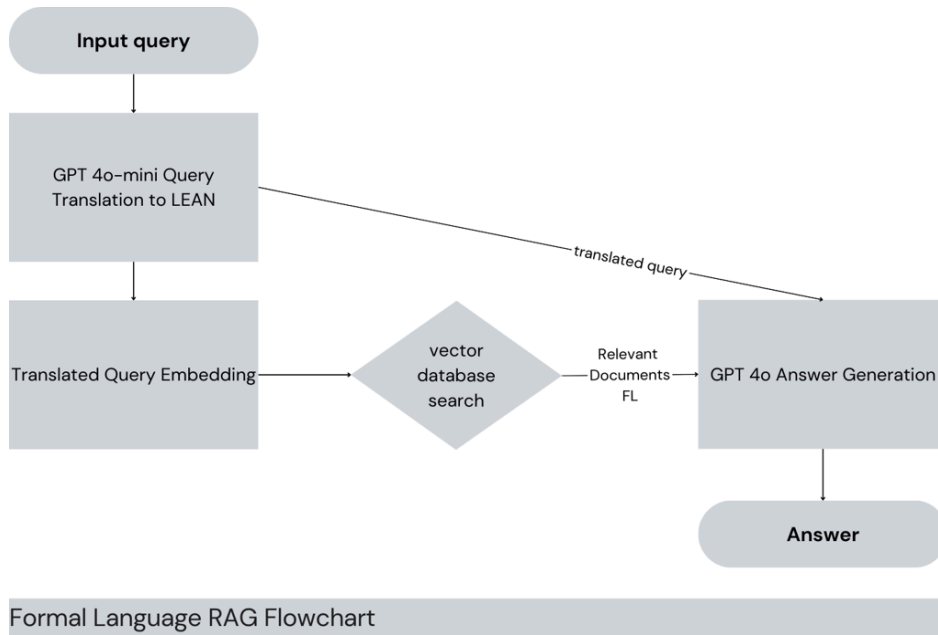

Formal Language RAG Flowchart

*C. Model Fine-Tuning*

The fine-tuning of GPT-4o-mini was a critical component of the FL RAG system. We trained the model on the dataset of NL and Lean code pairs, optimizing it to accurately translate NL mathematical statements into Lean code. The prompt used during fine-tuning was *"Translate statement into LEAN:"*. The training data used to fine tune the model are over 10 million tokens worth of statement pairs in NL and FL taken from the Lean Workbook dataset [Ying et al., 2024], each statement in natural language was paired by a translated statement in Lean, and signed with "by sorry" indicating the end of the proof.

The following is an example of a statement that we translated using the fine-tuned model:

**Example Statement:** Let h(b) = 444*b. Let r(l) = -l + 15369.

**Translated Output:** theorem lean_workbook_plus_28679 (b l : $\mathbb{R}$) (h : $\mathbb{R} \to \mathbb{R}$) (r : $\mathbb{R} \to \mathbb{R}$) (hh : h b = 444 * b) (hr : r l = -l + 15369) : h (r l) = 444 * (-l + 15369) := by sorry

## IV. EVALUATION

### A. Evaluation Method

In our study, we selected the Mathematics Dataset developed by Google [Saxton et al., 2019] as a primary benchmark to evaluate the performance of our systems. The Mathematics Dataset by Google is a large-scale collection of mathematical problems crafted to probe the mathematical understanding of machine learning models. It encompasses a wide range of mathematical domains, including arithmetic, algebra, calculus, probability, and number theory. Each problem is presented in NL, reflecting the way mathematical questions are typically posed in educational and real-world contexts. This dataset is particularly suitable for assessing mathematical reasoning capabilities in language models due to its comprehensive coverage of mathematical topics and its design to challenge models across various problem-solving scenarios.

One of the primary reasons for choosing this dataset is its diversity and depth. The problems vary in difficulty, ranging from simple computations to complex reasoning tasks that require multiple steps and a deep understanding of mathematical concepts. This variation enables a thorough assessment of our RAG systems' capabilities across different levels of complexity. It challenges the models not only in terms of computational skills but also in understanding problem statements, applying appropriate methods, and producing precise answers.

Moreover, the dataset's extensive range of topics ensures that our evaluation is comprehensive. By covering a broad spectrum of mathematical areas, we can test the models' proficiency in various domains, identifying strengths and weaknesses in their reasoning abilities. This comprehensive coverage is essential for demonstrating the effectiveness of integrating FL representations into RAGs, as it provides a robust testing ground for the models' enhanced reasoning capabilities.

The dataset's structure also provides practical benefits for our research. The problems are algorithmically generated with controlled variability, ensuring consistency in format while covering a vast array of question types. This consistency simplifies the preprocessing and integration of the dataset into our evaluation pipeline. Additionally, the availability of precise answers for each problem facilitates objective evaluation metrics, such as accuracy and correctness, allowing us to systematically measure the performance of our models. Furthermore, the NL format of the problems aligns well with our initial input modality, making the dataset particularly relevant for the NL RAG system. For the FL RAG, the dataset provides realistic scenarios for translating queries into formal representations, testing the effectiveness of our approach in practical contexts.

### B. Results

All of the questions used were taken from the hard section, we sampled 10 questions out of every category for 56 different categories of questions, which totals to 560 questions with answers being numeric, algebraic formulas, or true-or-false expressions. The NL RAG achieved an average of 54% answer correctness over all categories, the FL RAG achieved an average of 73% answer correctness, giving a difference in score of 19% over the NL RAG thus yielding a 35% performance boost.

Looking at some of the similarities is the performance of the models tested; they both seemed to perform well in the same sections (namely comparason_sort), and poorly on the same sections as well (namely algebra_sequence_nth_term), although the FL RAG out-performed the NL RAG in both cases. Both models handled questions with large numbers well when the computational tasks were simple (such as addition and multiplication), however the correctness suffered when dealing with more complicated computations (such as finding the gcd of two number or the prime factors of a number), however they both performed well in complex computations when the numbers were smaller and simpler.

Looking at some of the differences in the evaluation results; The FL RAG seemed to more significantly outperform the NL RAG in questions requiring symbolic reasoning, namely algebra and calculus questions. In the instances where the NL RAG got a wrong answer, it would simply output a wrong number whereas the FL RAG would sometimes claim that no answer could be found or give proof instructions that are irrelevant to the given questions, these mistakes are most likely a product of inaccuracies in the translation process.

*C. Examples*

The following are examples of questions from the test dataset along with their answers according to the models tested:

**1) Question:** Solve 312*s + 276*s - 661*s + 952 = -362 for s.
**Real Answer:** 18
**NL RAG Answer:** The final answer is {1}.
**FL RAG Answer:** The final answer is {18}.

**2) Question:** In base 10, what is -15 - -839090?
**Real Answer:** 839075
**NL RAG Answer:** The final answer is {-839075}.
**FL RAG Answer:** Therefore, the final answer is {839075}.

**3) Question:** What is the difference between -988.36 and -73357.6?
**Real Answer:** 72369.24
**NL RAG Answer:** The final answer is {72369.24}.
**FL RAG Answer:** It seems like you might have entered a statement instead of a math problem. Could you please provide the math problem you would like me to solve?

**4) Question:** Find the second derivative of 10490087*j*z**2 - 4460*j - 179*z wrt z.
**Real Answer:** 20980174*j
**NL RAG Answer:** The final answer is ({2 cdot 10490087 cdot j}).
**FL RAG Answer:** The final answer is {20980174 * j}.

**5) Question:** Suppose 27*t = 43 + 11. Let f be (t + (-111)/(-12))/((-2)/(-8)). What is the highest common divisor of f and 15?
**Real Answer:** 15
**NL RAG Answer:** Thus, the final answer is {15}.
**FL RAG Answer:** The final answer is {3}.

We can see in examples 1 and 5 instances where the NL RAG was simply more accurate. In example 4 we see an instance where both answered correctly, but the NL RAG didn't quite give a final answer. In example 2 we see a trick question where the NL RAG was tricked by but the FL RAG wasn't, probably due to the extra layer of formalization. In example 3 we see an instance where the NL RAG answered correctly, but the FL RAG didn't answer at all, probably an issue with the translation to FL.

## V. Conclusion

This study investigated the integration of formal mathematical language, specifically Lean code, into RAGs to enhance the mathematical reasoning capabilities of LLMs. By constructing a knowledge corpus composed of formalized mathematical statements and proofs, and translating NL queries into formal representations, we aimed to improve LLM performance in advanced logical reasoning tasks. The evaluation conducted using the Mathematics Dataset developed by Google [Saxton et al., 2019] demonstrated that the FL RAG achieved an average answer correctness of 73%, outperforming the NL RAG system, which achieved 54%. This significant improvement suggests that incorporating FL into the RAG process can yield advantages over conventional NL-based approaches in mathematical problem-solving. Notably, the knowledge corpus used in the FL RAG consisted of the same NL statements from the NL RAG translated into Lean. Therefore, even with potential mistakes and inaccuracies in the translation process—both when translating the query in real time and in creating the FL knowledge corpus—these improvements were achieved.

## VI. Discussion

While we can only speculate about the underlying reasons for the performance boost observed with the FL configuration, several factors may contribute to this improvement. One possible reason is the additional processing involved in the translation step. In many instances, the Lean translator was observed to attempt to solve the questions by providing a solution as part of the statement (sometimes incorrectly). This could inadvertently provide the LLM with more context or hints toward the correct answer.

Another potential factor is that the use of Lean in querying the LLM may help it access more mathematically accurate portions of its training data. Since data written in Lean is likely to be a more precise and rigorous source of mathematical knowledge than data in other languages, incorporating Lean code could enhance the model's reasoning capabilities. This suggests that our approach may serve as the basis for a novel inference technique that leverages formal language representations.

Finally, the performance boost could simply be due to the fact that using formal language is more effective for communicating mathematical ideas. Formal languages eliminate ambiguities inherent in natural language, providing clear and unambiguous representations of mathematical concepts. This clarity may give us an edge in training and instructing LLMs for the purposes of performing mathematical reasoning.

To gain a deeper understanding of these factors, further evaluation over more tests and tasks is necessary. Future work could involve testing the proof-writing capabilities of the models and their ability to answer more advanced questions that require larger context sizes, the use of planning, and more methodical approaches to execution. There is ample room for improvement in terms of the architectures used, system prompts, instruction methods, and data selection. Exploring these avenues could help in finding optimal configurations to further enhance the performance of LLMs in advanced logical reasoning tasks as discussed in this research.

## REFERENCES

Yuhuai Wu, Albert Qiaochu Jiang, Wenda Li, Markus Rabe, Charles Staats, Mateja Jamnik, Christian Szegedy. Autoformalization with Large Language Models 2022

Huajian Xin, Daya Guo, Zhihong Shao, Zhizhou Ren, Qihao Zhu, Bo Liu, Chong Ruan, Wenda Li, Xiaodan Liang. DeepSeek-Prover: Advancing Theorem Proving in LLMs through Large-Scale Synthetic Data 2024

Ayush Agrawal, Siddhartha Gadgil, Navin Goyal, Ashvni Narayanan, Anand Tadipatri. Towards a Mathematics Formalisation Assistant using Large Language Models 2024

Haohan Lin, Zhiqing Sun, Yiming Yang, Sean Welleck. Lean-STaR: Learning to Interleave Thinking and Proving 2024

Huaiyuan Ying, Zijian Wu, Yihan Geng, Jiayu Wang, Dahua Lin, Kai Chen. Lean Workbook: A large-scale Lean problem set formalized from natural language math problems 2024

Grégoire Mialon, Roberto Dessì, Maria Lomeli, Christoforos Nalmpantis, Ram Pasunuru, Roberta Raileanu, Baptiste Rozière, Timo Schick, Jane Dwivedi-Yu, Asli Celikyilmaz, Edouard Grave, Yann LeCun, Thomas Scialom. Augmented Language Models: a Survey 2023

Yunfan Gao, Yun Xiong, Xinyu Gao, Kangxiang Jia, Jinliu Pan, Yuxi Bi, Yi Dai, Jiawei Sun, Meng Wang, Haofen Wang. Retrieval-Augmented Generation for Large Language Models: A Survey. 2023

Aitor Lewkowycz, Anders Andreassen, David Dohan, Ethan Dyer, Henryk Michalewski, Vinay Ramasesh, Ambrose Slone, Cem Anil, Imanol Schlag, Theo Gutman-Solo, Yuhuai Wu, Behnam Neyshabur, Guy Gur-Ari, Vedant Misra. Solving Quantitative Reasoning Problems with Language Models. 2022

David Saxton, Edward Grefenstette, Felix Hill, Pushmeet Kohli. Analysing Mathematical Reasoning Abilities of Neural Models. 2019